\documentclass[11pt]{article}

\usepackage[final]{acl}

\usepackage{times}
\usepackage{latexsym}
\usepackage{booktabs}
\usepackage{multirow}
\usepackage{tcolorbox}
\usepackage{tabularx}
\tcbuselibrary{listings, breakable}

\usepackage[T1]{fontenc}

\usepackage[utf8]{inputenc}

\usepackage{microtype}

\usepackage{inconsolata}

\usepackage{graphicx}

\title{Prompt-Level Distillation: A Non-Parametric Alternative to Model Fine-Tuning for Efficient Reasoning}

\author{Sanket Badhe\thanks{\ \ Equal contribution.} \\
  Google \\
  Mountain View, California, USA \\
  \texttt{sanketbadhe@google.com} \\\And
  Deep Shah\footnotemark[1]\\
  Google \\
  Mountain View, California, USA \\
  \texttt{shahdeep@google.com}}

\begin{document}
\maketitle
\begin{abstract}
Advanced reasoning typically requires Chain-of-Thought prompting, which is accurate but incurs prohibitive latency and substantial test-time inference costs. The standard alternative, fine-tuning smaller models, often sacrifices interpretability while introducing significant resource and operational overhead. To address these limitations, we introduce Prompt-Level Distillation (PLD). We extract explicit reasoning patterns from a Teacher model and organize them into a structured list of expressive instructions for the Student model's System Prompt. Evaluated using Gemma-3 4B, PLD improved Macro F1 scores on StereoSet (57\% to 90.0\%) and Contract-NLI (67\% to 83\%), while increasing LogiQA accuracy to 70\%. Similar results on Mistral Small 3.1 demonstrate cross-architecture generalizability, enabling these compact models to match frontier performance with negligible latency overhead. These expressive instructions render the decision-making process transparent, allowing for full human verification of logic, making this approach ideal for regulated industries such as law, finance, and content moderation, as well as high-volume use cases and edge devices.
\end{abstract}

\section{Introduction}

Large Language Models (LLMs) have established themselves as the defacto standard for complex reasoning tasks in industry, ranging from financial analysis to automated code generation \cite{comanici2025gemini25pushingfrontier, openai2023gpt4}. This capability is largely underpinned by Chain-of-Thought (CoT) prompting, which elicits multi-step reasoning by encouraging models to generate intermediate rationales before producing a final answer \cite{wei2022chain, kojima2022large}. Further refinements, such as Self-Consistency \cite{wang2023self} and Least-to-Most prompting \cite{zhou2023least}, have demonstrated that explicitly structuring the reasoning process significantly improves accuracy on symbolic and arithmetic benchmarks. Consequently, CoT has become the default inference strategy for high-stakes and reasoning intensive applications.

However, this accuracy gain incurs a latency penalty that is often prohibitive for real-time production environments. CoT maximizes performance by generating verbose reasoning traces, often hundreds of tokens long, which linearly increase inference latency and computational cost \cite{meincke2025decreasing}. In addition, human-like reasoning capabilities only emerges in language models with large number of parameters \cite{webb2023emergent} making it challenging to use smaller models out of the box. In high-throughput settings, this creates a sharp conflict: practitioners are forced to choose between high-accuracy models that utilize extensive test-time compute \cite{nye2021show} or efficient zero-shot inferences that sacrifice reasoning depth for speed \cite{brown2020language, pope2022efficiently}.

The standard industry solution to this efficiency bottleneck is Knowledge Distillation (KD), where a smaller student model is fine-tuned to mimic the outputs of a larger teacher \cite{hinton2015distilling, hsieh2023distilling}. Recent approaches, such as Distilling Step-by-Step \cite{hsieh2023distilling} and Fine-tune-CoT \cite{ho2023large}, have successfully transferred reasoning capabilities to smaller architectures \cite{fu2023specializing, magister2023teaching}. Yet, traditional KD presents significant operational flaws for modern industry stacks. First, attempting to fine-tune smaller models purely on the text outputs of proprietary APIs has been shown to successfully imitate style, but broadly fails to transfer underlying reasoning capabilities \cite{gudibande2023false}. Second, it introduces maintenance debt; every time the teacher model improves or the domain logic shifts, the student must be retrained, requiring the management of diverse model artifacts \cite{gou2021knowledge}. Third, in practice, maintaining a fine-tuned model for every use-case is not scalable, since it directly impacts the resource allocation and maintenance cost. Furthermore, relying on complex external retrieval frameworks introduces its own set of paradigms and operational pitfalls \cite{shah2026taxonomy}. Finally, fine-tuning often requires thousands of curated examples to avoid catastrophic forgetting or hallucination, making it resource-intensive for agile teams \cite{luo2025empirical} and it's difficult collect huge data for Long-Tail Knowledge \cite{badhe2026long}.

To address these limitations, we introduce Prompt-Level Distillation (PLD), a novel framework that uses a labeled training dataset to transfer the reasoning skills of a large teacher model into the system prompt of a smaller student model. This allows the student to achieve the accuracy of Chain-of-Thought reasoning with the speed of zero-shot inference, all without updating any model parameters. Unlike traditional distillation that compresses knowledge into weights, PLD transfers reasoning by compiling the Teacher's logic directly into the System Prompt. We extract the explicit decision rules from individual training examples and consolidate them into a unified, conflict-free instruction set. While this framework is broadly applicable, we specifically focus on reasoning-intensive classification tasks, where the model must navigate complex logical constraints to reach a verifiable decision. By externalizing the reasoning process into a structured prompt, PLD enables a Student model to execute complex logic zero-shot, bypassing the need for generating intermediate reasoning tokens at runtime.

Our contributions are following:
\begin{itemize}
    \item \textbf{Framework:} We introduce Prompt-Level Distillation (PLD), a non-parametric alternative to fine-tuning that transfers reasoning capabilities by compiling Teacher logic into the Student's system prompt, thereby avoiding maintenance debt and parameter updates.
    \item \textbf{Methodology:} We propose a modular pipeline that combines supervised instruction extraction, clustering, and a novel closed-loop Conflict Resolution phase to synthesize robust, contradiction-free reasoning heuristics.
    \item \textbf{Performance:} We demonstrate that PLD enables compact models (e.g., Gemma-3 4B, Mistral Small 3.1) to match frontier-level reasoning with the low latency of zero-shot inference, effectively decoupling reasoning depth from computational cost.
\end{itemize}

\section{Related Work}

\subsection{Chain-of-Thought and Inference Efficiency}
CoT prompting has established itself as the primary method for eliciting multi-step reasoning in LLMs \cite{wei2022chain, kojima2022large}. By decomposing complex problems into intermediate steps, CoT significantly improves performance on symbolic and arithmetic benchmarks \cite{wang2023self, zhou2023least, yao2023tree}. However, this accuracy imposes substantial inference latency and computational overhead, as models must autoregressively generate lengthy rationales for every query \cite{nye2021show, meincke2025decreasing}. 

Recent efforts to mitigate this latency bottleneck have focused on condensing the reasoning process. Approaches such as Conditioned Compressed CoT (C3oT) \cite{kang2025c3ot}, CROP: Token-Efficient Reasoning \cite{shah2026crop} and Focused Chain-of-Thought \cite{struppek2025focused} attempt to shorten reasoning traces without losing information. Others, like SpecCoT \cite{shi2025speccot}, employ speculative decoding frameworks to accelerate the generation of reasoning steps via smaller draft models. While these methods reduce marginal token costs, they fundamentally retain the runtime computation of reasoning. In contrast, our PLD framework effectively computes reasoning offline. By identifying and clustering valid reasoning paths during a preprocessing phase, we cache the resulting logic into the system prompt, allowing the student model to execute complex heuristics instantaneously without generating intermediate tokens.

\subsection{Knowledge Distillation and Reasoning Transfer}
Knowledge Distillation (KD) traditionally serves as the primary mechanism for transferring capabilities from capable teacher models to efficient students \cite{hinton2015distilling}. In the context of LLMs, this typically involves minimizing the Kullback-Leibler divergence between the teacher's and student's logic distributions \cite{hsieh2023distilling, magister2023teaching}. Recent advancements have extended this to opaque settings where logits are inaccessible, relying instead on fine-tuning students on generated samples \cite{ye2025black, li2025learning}. For instance, Fine-tune-CoT \cite{ho2023large} and Distilling Step-by-Step \cite{hsieh2023distilling} demonstrate that smaller models can internalize reasoning patterns through supervised training on teacher traces.

Despite their efficacy, these methods remain parametric: they require weight updates, extensive training corpora, and rigorous hyperparameter tuning to avoid catastrophic forgetting \cite{luo2025empirical}. Our approach diverges by proposing a non-parametric distillation method. Instead of compressing intelligence into model weights, PLD distills intelligence into the context window. This allows for the transfer of reasoning capabilities to closed-weight models purely through prompt injection, eliminating the maintenance debt associated with managing fine-tuned model artifacts. Crucially, this externalization converts the black-box decision boundaries of fine-tuning into transparent, natural language instructions, enabling full human verification of logic, a capability typically lost in standard parametric distillation.

\subsection{Automatic Prompt Optimization (APO)}
The rapid evolution of LLMs has necessitated automated techniques for prompt engineering, broadly categorized as Automatic Prompt Optimization (APO) \cite{ramnath2025systematic}. Seminal works like Automatic Prompt Engineer (APE) \cite{zhou2022large} and Optimization by PROmpting (OPRO) \cite{yang2023large} treat prompt generation as a evolutionary search problem, utilizing LLMs to iteratively propose and score instruction phrasing. Closely related is the field of Instruction Induction, where models often derive a single task descriptions from input-output pairs \cite{honovich2023instruction}. More structured frameworks, such as DSPy \cite{khattab2023dspy} and TextGrad \cite{yuksekgonul2024textgrad}, optimize multi-stage pipelines by treating prompts as modular programs or applying textual gradients to refine system components based on feedback. Existing surveys categorize these methods by their optimization surface, ranging from discrete text instructions \cite{pryzant2023automatic, guo2024connecting} to continuous soft prompts \cite{lester2021power, li-liang-2021-prefix}.

However, the primary objective of existing APO methods is instructional refinement—finding the optimal wording to align a model with a task \cite{cheng-etal-2024-black, fernando2023promptbreeder1, prasad-etal-2023-grips}. They largely treat the reasoning logic as implicit within the model. Furthermore, unlike Prompt Compression techniques that reduce latency by pruning low-information tokens \cite{jiang2023llmlingua}, PLD performs semantic compression. It effectively offloads the computational cost of reasoning from runtime generation to offline compilation. PLD differs fundamentally in its objective: it is not searching for better phrasing or fewer tokens, but rather synthesizing logical heuristics. Through our Supervised Instruction Extraction and Clustering Logic Synthesis pipeline, we explicitly mine, cluster, and codify the domain-specific heuristics used by the teacher model. The result is not just a better-worded instruction, but a portable library of reasoning patterns injected into the student's system prompt.

\section{Methodology}
\label{sec:methodology}

\begin{figure*}[ht] 
    \centering
    \includegraphics[width=0.95\textwidth]{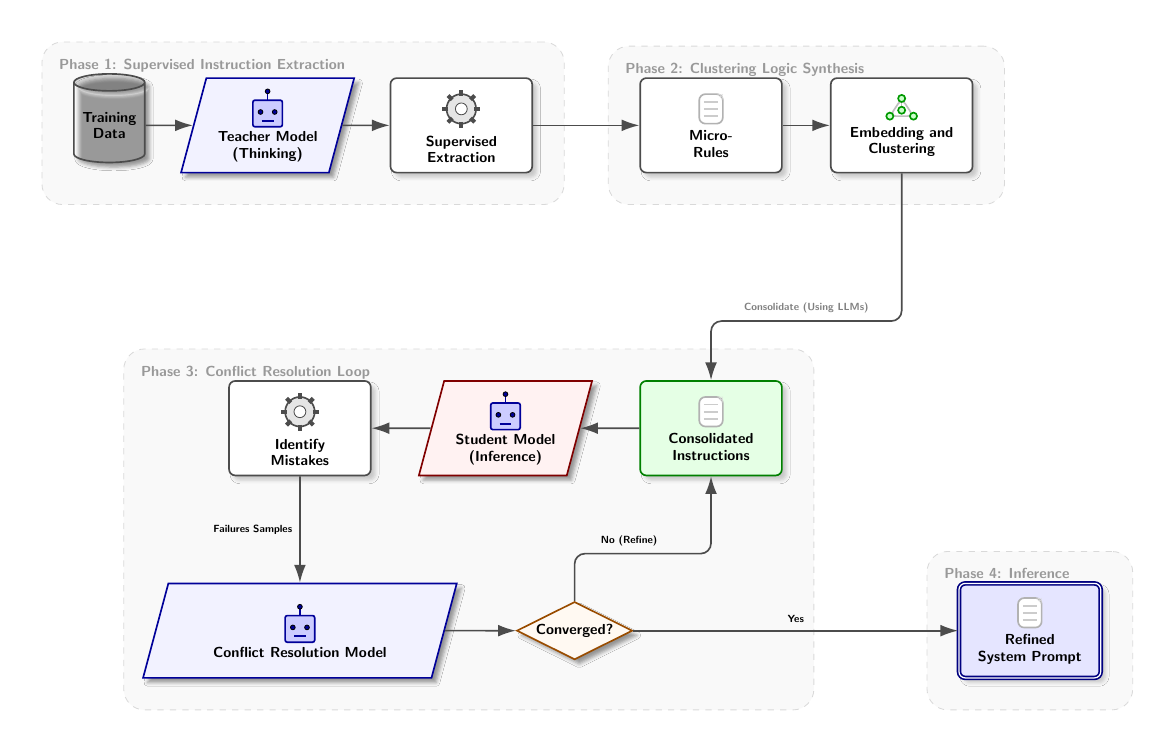}
    \caption{\textbf{Overview of Prompt-Level Distillation (PLD).} The framework operates in four phases: (1) Supervised Instruction Extraction from training data, (2) Semantic Synthesis using clustering, (3) A Closed-Loop Conflict Resolution phase to refine logic, and (4) Zero-shot Inference using the consolidated system prompt.}
    \label{fig:framework}
\end{figure*}

We propose \textit{Prompt-Level Distillation} (PLD), a framework for transferring reasoning capabilities from reasoning-optimized models to efficient inference models without parameter updates. In this supervised setting, we treat the System Prompt as a comprehensive set of reasoning instructions that is mined from a labeled training dataset, synthesized to remove redundancy, and rigorously validated. As illustrated in Figure~\ref{fig:framework}, the framework operates through four distinct phases: (1) Supervised Instruction Extraction from training samples, (2) Semantic Instruction Synthesis, (3) Closed-Loop Conflict Resolution, and (4) Inference.

\subsection{Phase 1: Supervised Instruction Extraction}
\label{subsec:elicitation}

To generate high-quality training data, we leverage the self-reflective capabilities of reasoning-optimized models. We employ a strategy that generates reasoning trace and extracts them into instruction in a single inference step.

Given a labeled training set $T = \{(x_i, y_i)\}$, we construct a structured prompt that compels the teacher model to perform two tasks simultaneously:


\begin{enumerate}
    \item \textbf{Supervised Problem Solving:} The model employs chain-of-thought reasoning to analyze the logical constraints of the input $x_i$ and rationalize the provided ground-truth label $y_i$.
    \item \textbf{Instruction Abstraction:} The model immediately abstracts this reasoning process into a generalized natural language instruction, removing specific entity names while preserving the causal mechanism required to reach the ground truth.
\end{enumerate}

This results in an augmented dataset $D = \{(x_i, y_i, I_i)\}$, where $I_i$ is the abstract instruction derived from the training example. This approach reduces the computational overhead of data generation by eliminating the need for multi-stage pipelines.

\subsection{Phase 2: Clustering Logic Synthesis}
\label{subsec:synthesis}

Raw extracted instructions are often redundant or overly specific to individual training examples. To create a generalized instruction set, we employ DBSCAN\cite{ester1996density} to group these instructions by semantic similarity. Unlike partition-based methods (like K-Means) that force every data point into a cluster, DBSCAN explicitly isolates noise points, ensuring that our final prompts are not polluted with non-generalizable instructions.

We represent each micro-instruction as a dense vector using a high-fidelity embedding model \cite{lee2025geminiembeddinggeneralizableembeddings}. We then apply DBSCAN using a cosine distance metric to identify dense regions of reasoning patterns. This approach offers a critical advantage: it allows the natural number of logical rules to emerge from the data density without requiring us to specify the number of clusters. 

A model (same as the teacher model) processes each identifiable dense cluster to synthesize the constituent micro-rules into a single, unified instruction. Outlier points identified as noise by DBSCAN are discarded.

\subsection{Phase 3: Conflict Resolution}
\label{subsec:refinement}

Conflicting instructions from Phase 1 would be randomly merged in the previous summarization phase leading to suboptimal instructions. To ensure reliability, we implement a conflict resolution loop that iterates over the instructions, identify gaps against training data, eventually refining instructions.

\begin{enumerate}
    \item \textbf{Inference and Error Analysis:} We deploy the student model with the current consolidated instruction set to perform inference on the given training data set. We explicitly isolate mistake examples where the model followed the instruction but failed to match the ground-truth label.
    \item \textbf{Adversarial Refinement:} These failures along with successful examples are sampled and fed into a Conflict Resolution Model (same as teacher model). The model analyzes the root cause and generates an updated instruction that learns better prompt improving accuracy. Providing correct examples is critical to avoid performance degradation.
    \item \textbf{Convergence:} This cycle repeats until the error rate on the validation set converges
\end{enumerate}

\subsection{Phase 4: Inference}
\label{subsec:inference}

For deployment, the refined system prompt is injected into the student model. We provide the complete set of consolidated instructions to the model, enabling it to handle the full range of logical constraints defined in the training data without the latency overhead of external retrieval mechanisms.

\begin{table*}[!ht]
\centering
\begin{tabular}{l cccc}
\toprule
\textbf{Method} & \textbf{Gemma} & \textbf{Gemini 2 Flash} & \textbf{Gemini 3 Flash} & \textbf{Mistral Small 3.1} \\
\midrule
\multicolumn{5}{c}{\textbf{Stereoset (Macro-F1)}} \\
\midrule
Zero-shot & 0.57 & 0.53 & 0.92 & 0.65 \\
Few-shot & 0.75 & 0.83 & 0.90 & 0.77 \\
Fine-tuning & 0.89 & - & - & 0.96 \\
TextGrad & 0.87 & 0.90 & 0.91 & 0.96 \\
Clustered-Inst. (Our) & 0.90 & 0.92 & 0.93 & 0.96 \\
Post-Conflict (Our) & 0.90 & 0.93 & 0.93 & 0.97 \\
\midrule
\multicolumn{5}{c}{\textbf{Contract-NLI (Macro-F1)}} \\
\midrule
Zero-shot & 0.67 & 0.73 & 0.77 & 0.71 \\
Few-shot & 0.70 & 0.75 & 0.79 & 0.71 \\
Fine-tuning & 0.76 & - & - & 0.77 \\
TextGrad & 0.74 & 0.77 & 0.76 & 0.73 \\
Clustered-Inst. (Our) & 0.81 & 0.80 & 0.84 & 0.75 \\
Post-Conflict (Our) & 0.83 & 0.83 & 0.86 & 0.78 \\
\midrule
\multicolumn{5}{c}{\textbf{LogiQA (Accuracy)}} \\
\midrule
Zero-shot & 0.67 & 0.64 & 0.80 & 0.56 \\
Few-shot & 0.66 & 0.64 & 0.81 & 0.55 \\
Fine-tuning & 0.67 & - & - & 0.55 \\
TextGrad & 0.69 & 0.66 & 0.84 & 0.56 \\
Clustered-Inst. (Our) & 0.69 & 0.67 & 0.84 & 0.57 \\
Post-Conflict (Our) & 0.70 & 0.67 & 0.84 & 0.56 \\
\bottomrule
\end{tabular}
\caption{Performance comparison on StereoSet, Contract-NLI and LogiQA  datasets using Gemma, Gemini 2 Flash, Mistral Small 3.1, and Gemini 3 Flash (Teacher).}
\label{tab:dataset_results}
\end{table*}

\section{Experimental Setup}

\subsection{Setup}

For the supervised instruction extraction phase, we employed Gemini 3 Flash, configured with its thinking mode as our teacher model. Given the complexity of synthesizing consolidated instructions and conflict resolution phase, we required a highly capable model; thus, we opted for Gemini 3 Pro, also leveraging its thinking mode.

We clustered the extracted instructions using DBSCAN on Gemini Embedding vectors \cite{lee2025geminiembeddinggeneralizableembeddings} and synthesized the resulting clusters into unified heuristics using Gemini 3 Pro, with full hyperparameters and prompt details provided in Appendix~\ref{app:implementation_details}.

We evaluated the efficacy of our PLD framework using distinct student models of various size and families: Gemma-3 4B \cite{team2025gemma}, Mistral Small 3.1 24B \cite{liu2026ministral} and Gemini 2 Flash \cite{comanici2025gemini}. Gemma-3 4B and Mistral Small 3.1 24B was selected due to its compact parameter footprint and highly efficient inference. Similarly, Gemini 2 Flash was chosen because it offers significantly lower inference cost compared to current frontier models (such as Gemini 3 Pro and Gemini 3 Flash), making both models ideal candidates for assessing reasoning transfer to differently resource-constrained environments.

\subsection{Baseline}


We evaluate PLD against standard Zero-shot and Few-shot ($k=5$, randomly sampled) prompting, as well as TextGrad \cite{yuksekgonul2024textgrad}, a state-of-the-art framework for APO. To benchmark against parametric approaches, we include results for LoRA \cite{hu2022lora} fine-tuning the Gemma-3 4B student model on teacher-generated reasoning traces. Furthermore, to isolate and quantify the specific impact of the closed-loop conflict resolution phase, we introduce an ablation setting that evaluates the performance of the intermediate instructions extracted immediately prior to this refinement step.

\subsection{Datasets}

We evaluate our proposed Prompt-Level Distillation framework on two distinct datasets, representing varying degrees of reasoning complexity:

\textbf{1. Contract NLI} \cite{koreeda2021contractnli}: This dataset consists of contract-hypothesis pairs where the objective is to predict the logical relationship between the document premise and the hypothesis across three classes: \textit{Entailment}, \textit{Contradiction}, and \textit{NotMentioned}. We selected this dataset because legal domain classification demands rigorous logical deduction, requiring models to navigate complex linguistic structures and nuanced negations.

\textbf{2. StereoSet} \cite{nadeem2021stereoset}: Designed to measure stereotypic biases learned by language models across four domains (Gender, Race, Profession, and Religion), we are using this dataset to assess our methods effectiveness on relatively simple task. The model is tasked with predicting the correct underlying domain given an isolated context sentence.

\textbf{3. LogiQA} \cite{liu2020logiqa}: A challenging machine reading comprehension dataset sourced from expert logical examinations. It requires complex deductive inference (e.g., categorical, conditional, and disjunctive reasoning), assessing PLD's ability to successfully extracts and generalizes intricate reasoning chains for logical deduction tasks. 

\begin{table}[h]
\centering
\resizebox{\columnwidth}{!}{
\begin{tabular}{lcccc}
\hline
Dataset & Train & Validation & Test & Total \\ \hline
Contract NLI & 7,191 & 1,037 & 2,091 & 10,319 \\
Stereoset & 842 & 211 & 1,053 & 2,106 \\
LogiQA & 7376 & 651 & 651 & 8,678 \\
\hline
\end{tabular}
}
\caption{Dataset Split Statistics}
\label{tab:dataset_splits}
\end{table}

\section{Results and Analysis}

\subsection{Overall Performance}


Table~\ref{tab:dataset_results} details the performance of our Prompt-Level Distillation (PLD) framework against standard zero-shot and few-shot baselines. Across StereoSet, Contract-NLI, and LogiQA, PLD consistently yields the better performance against baseline, validating the efficacy of non-parametric reasoning transfer. 

Our method showed strong performance gains on Gemma-3 4B, which is 25 times cheaper (Appendix \ref{sec:model_pricing}) and ~80 times faster (Appendix \ref{sec:model_latency}) than Gemini-3 Flash. On the StereoSet task, applying the fully refined PLD prompt drives macro-F1 to 0.90, a +0.33 point absolute improvement over zero-shot prompting. On the logically demanding Contract-NLI dataset, Gemma-3 4B achieves an 0.83 macro-F1, decisively outperforming its zero-shot baseline of 0.67. Furthermore, on LogiQA, PLD improves the accuracy of both Gemma-3 4B (from 0.67 to 0.70) and Gemini 2 Flash (from 0.64 to 0.67), demonstrating that the framework effectively boosts performance on benchmarks requiring multi-step logical deduction. Notably, Mistral Small 3.1 (24B) exhibited similar performance trends, confirming that PLD-distilled instructions transfer successfully across diverse model architectures. Our method also benefits the teacher model (Gemini 3 Flash) by explicitly distilling reasoning, providing a macro-F1 score of 0.86 on Contract-NLI compared to its 0.77 zero-shot baseline.


\subsection{Analysis}
\textbf{Disproportionate Gains for Compact Models:} Zero-shot and few-shot baselines reveal a significant performance gap between the powerful Gemini 3 Flash and more compact architectures. However, applying PLD effectively closes this gap. Both Gemma-3 4B and Mistral Small 3.1 achieve performance parity with larger frontier models after distillation, highlighting the framework's efficacy in transferring reasoning capabilities across diverse, resource-constrained architectures.

\textbf{Crucial Role of Conflict Resolution in Complex Domains:} The closed-loop conflict resolution phase yielded a 2.5\% increment macro-F1 improvement on the structurally complex Contract-NLI dataset. Initial single-example extractions seldom generates contradictory instructions often missing broader linguistic nuances. The refinement loop successfully rectified these contradictions by learning various nuances previously missed. Conversely, this phase provided negligible gains on the StereoSet, indicating that iterative refinement is primarily essential for tasks involving intricate and overlapping edge cases. Conflict resolution loop converged on first iteration on Stereoset and second iteration on Contract-NLI.

\textbf{Preserving Minority and Edge Cases:} 
The teacher model tasked with synthesizing clustered instructions often discards conflicting rules associated with minority labels and edge cases (see Appendix~\ref{sec:appendix_conflicts_resolved_examples}). Since the supervised extraction operates on isolated examples, it inherently produces narrow, sometimes conflicting heuristics (Appendix~\ref{sec:appendix_single_extraction_conflicts}), often due to not picking up the correct nuances in the example at first. The Conflict resolution Loop is therefore critical not just for overall accuracy, but for preserving the logical coverage of minority constraints within the consolidated prompt.

\textbf{Exhaustive Reasoning Extraction vs. Implicit Parametric Reasoning:} The superior performance of PLD over TextGrad and Fine-tuning, particularly on Contract-NLI dataset highlights the advantage of externalizing logic rather than relying on latent model capacity. While \textit{Fine-tuning} attempts to embed reasoning patterns directly into a model’s weights, internalizing a truly comprehensive and exhaustive set of instructions is often infeasible for compact models with limited examples; often struggling to preserve high-precision rules for long-tail edge cases. Furthermore, while frameworks like \textit{TextGrad} optimize prompt phrasing to better elicit existing capabilities, they still fundamentally rely on the student model to think through constraints at inference. In contrast, PLD externalizes this reasoning by providing pre-mined logical heuristics directly in the prompt. By capturing complex nuances and contradictions within the instruction set itself, PLD minimizes the cognitive load on the student, enabling it to execute sophisticated logic that it could not otherwise derive independently.

\section{Conclusion}
In this work, we introduced Prompt-Level distillation, a simple and general-purpose framework for the automatic optimization of LLM prompts. We employ a novel technique which distills the per-example instructions by teacher into the prompt as opposed to fine-tuning model. We evaluated our method against two benchmarks and suggests that our method significantly improves the performance without any fine-tuning or model training. We also observed that our conflict resolution loop provides the most value on complex tasks involving tricky edge cases.

We believe our work can be carried forward in various directions. One can experiment with smarter solutions when sampling failures in conflict resolution loop. We deployed embedding based solution for clustering, but given the rise of In-context clustering \cite{wang2025context} and increase in model context window, once can consider deploying it to generate more cohesive clusters which would directly affect the generated consolidated instructions. 

\section{Limitations}
We focus our current evaluation on reasoning-intensive classification tasks where the model must navigate complex, static decision boundaries (e.g., regulatory compliance). Consequently, PLD may face limitations on tasks requiring dynamic, run-time computation (e.g., complex arithmetic or symbolic proofs), where reasoning cannot be fully externalized into a concise summary but instead requires generating intermediate tokens. Furthermore, we do not explicitly model the scaling limits of the system prompt; as task complexity increases, the consolidated instruction set may grow to exceed the effective context window or induce prompt-processing latency, potentially necessitating further compression techniques.

\section{Ethical Considerations}
Our evaluation includes the StereoSet dataset, which is designed to measure stereotypic biases. While our method effectively extracts classification logic, there is a risk that the teacher model may hallucinate or amplify biases present in the training data during the instruction extraction phase.  Practitioners must remain vigilant that the Consolidated Instruction Set does not encode discriminatory logic, as the Student model will follow these instructions faithfully.

\bibliography{custom}

\onecolumn
\appendix

\section{Learned instructions}
\label{sec:learned_instructions}

\begin{tcolorbox}[
    title=Instructions learned for Stereoset,
    colback=gray!5!white,
    colframe=gray!80!black
]
Identify the noun immediately preceding or modifying the 'BLANK' in the sentence and assign the category that matches it best: \\
- If it is a recognized job title, career, trade, or institutional role, categorize as profession. \\
- If it is a country, nationality, geographic origin, or demonym, categorize as race. \\
- If it is a gender-specific pronoun, kinship term, or explicitly gendered word, categorize as gender. \\
- If it is a religious group, caste, or sacred text, categorize as religion.
\end{tcolorbox}

\begin{tcblisting}{
    title=Subset of Instructions learned for Contract-NLI,
    colback=gray!5!white,
    colframe=gray!80!black,
    listing only,
    breakable
}
{
    "topic": "Retaining Copies (Archival)",
    "logic": [
      {
        "condition": "The text explicitly permits the Receiving Party to use certain information or copies for specific reasons (e.g., 'may retain', 'permitted to keep', 'exception to the return/destruction obligation') for purposes such as 'archival', 'legal', 'compliance', 'backup', or 'litigation' purposes.",
        "label": "Entailment"
      },
      {
        "condition": "The text explicitly prohibits the retention of any information or copies (keywords: 'retain no', 'not retain any', 'without retaining', 'no information... shall be retained') OR mandates that ALL materials and copies be returned or destroyed automatically with NO exceptions or archival rights mentioned.",
        "label": "Contradiction"
      },
      {
        "condition": "The text is silent regarding retention permissions, OR it mandates the return or destruction of information only 'upon request' or 'upon notice' without using strong specific negative language like 'shall not retain' or 'retaining no'.",
        "label": "NotMentioned"
      }
    ]
  }
  ...
  {
    "topic": "Reverse Engineering",
    "logic": [
      {
        "condition": "The text explicitly prohibits the act (keywords: reverse engineer, decompile, disassemble, reconstruct, reverse assemble, reverse compile, discover source code, analyze chemical structure, determine chemical composition, recreate material) without providing an exception that allows the act",
        "label": "Entailment"
      },
      {
        "condition": "The text explicitly permits reverse engineering or related activities under certain conditions or exceptions (e.g., 'except as needed for the project' or 'unless authorized'), which contradicts a blanket 'shall not' hypothesis",
        "label": "Contradiction"
      },
      {
        "condition": "The specific keywords (reverse engineer, decompile, disassemble, reconstruct, analyze chemical structure, recreate) are absent (even if the contract prohibits 'unauthorized use' or 'copying')",
        "label": "NotMentioned"
      }
    ]
  }
  ...
\end{tcblisting}

\subsection{Supervised Instruction Extraction Prompts}

\begin{tcolorbox}[
    title=Contract NLI Instruction Extraction Prompt,
    colback=gray!5!white,
    colframe=gray!80!black,
    breakable
]
\# Context \& Goal \\
You are the "Teacher" in a Logic Distillation pipeline. Your goal is to extract **executable logic rules** from a single training example. \\

\# Input Data \\
1.  **Contract Snippet:** A text from a legal agreement. \\
2.  **Hypothesis:** A claim about the contract. \\
3.  **Gold Label:** The correct classification (Entailment, Contradiction, NotMentioned). \\

\# Task \\
You must perform two tasks simultaneously to generate the output: \\
1. **Supervised Problem Solving:** Analyze the logical constraints of the input text and rationalize why the provided `gold\_label` is correct. \\
2. **Instruction Abstraction:** Immediately abstract this reasoning process into a generalized natural language instruction. Remove specific entity names but preserve the causal mechanism required to reach the ground truth. \\

\# Output Schema (JSON) \\
You must return a JSON object with the following fields: \\

\#\# 1. reasoning\_trace (String)
**Purpose:** The Chain-of-Thought rationale explaining why the label fits the text. \\
**Requirement:** Explicitly decompose the logical constraints found in the text. \\

\#\# 2. executable\_rule (String)
**Purpose:** This will be pasted into the System Prompt of a very small parameter model. \\
**Requirement:** Write a binary instruction. It must be: \\
    * **Binary:** Easy to check (Yes/No). \\
    * **Self-Contained:** Do not say "Think about the context." Say "Check if word X is present." \\
    * **Abstracted:** Generalize it slightly so it applies to similar contracts, not just this one. \\

contract\_snippet: <contract\_begin>\{sentence1\}<contract\_end> \\
hypothesis: <hypothesis\_begin>\{sentence2\}<hypothesis\_end> \\
gold\_label: \{gold\_label\} \\
\end{tcolorbox}

\begin{tcolorbox}[
    title=Stereoset Instruction Extraction Prompt,
    colback=gray!5!white,
    colframe=gray!80!black,
    breakable
]
\# Context \& Goal \\
You are the "Teacher" in a Logic Distillation pipeline. Your goal is to extract **executable logic rules** from a single training example. \\

\# Input Data \\
- Context: A sentence snippet containing a subject and a blank/placeholder. \\
- Gold Label: The bias\_type (e.g., profession, race, gender, religion). \\

\# Task \\
You must perform two tasks simultaneously to generate the output:
1. **Supervised Problem Solving:** Analyze the sentence structure and subject to rationalize why the `gold\_label` is correct.
2. **Instruction Abstraction:** Immediately abstract this reasoning process into a generalized natural language instruction. Remove specific entity names but preserve the logic required to identify the bias type. \\

\# Output Schema (JSON) \\
You must return a JSON object with the following fields: \\

\#\# 1. reasoning\_trace (String)
**Purpose:** The Chain-of-Thought rationale explaining why the label fits the context. \\
**Requirement:** Explicitly identify the semantic category of the subject (e.g., job title vs. religious group). \\

\#\# 2. executable\_rule (String)
**Purpose:** This will be pasted into the System Prompt of a very small parameter model. \\
**Requirement:** Write a binary instruction. It must be: \\
    * **Binary:** Easy to check (Yes/No). \\
    * **Self-Contained:** Do not say "Think about the context." Say "Check if the subject is X." \\
    * **Abstracted:** Generalize it so it applies to similar subjects, not just this specific word. \\
    * **Output:** Clearly state which category to assign if the rule is true. \\

\# Example to Process: \\
Context: \{context\} \\
Gold Label: \{bias\_type\} \\
\end{tcolorbox}

\section{Instruction Conflicts}
\subsection{Supervised Instruction Extraction conflicts}
\label{sec:appendix_single_extraction_conflicts}

Supervised Instruction Extraction can generate instructions which are conflicting with each other. Here are some of the examples of conflicting (A vs B) instruction we got from Contract-NLI dataset:

\begin{tcolorbox}[
    title=Conflicting (A vs B) Rules from Supervised Instruction Extraction (Contract-NLI),
    colback=gray!5!white,
    colframe=gray!80!black,
    fonttitle=\bfseries
]
\begin{tabularx}{\linewidth}{ 
  c | 
  >{\raggedright\arraybackslash}X | 
  >{\raggedright\arraybackslash}X 
}
\textbf{No.} & \textbf{Instruction A} & \textbf{Instruction B} \\
\hline
1 & If the contract states information must be returned "upon request," hypotheses claiming it must be returned "upon termination" are \textbf{Entailment}. & If the contract specifies return "upon request" but does not explicitly name "termination" as an automatic trigger, the claim is \textbf{NotMentioned}. \\
\hline
2 & General catch-all phrases like "in whatever form" or "of whatever nature" without the explicit spoken medium mean verbal information is \textbf{NotMentioned}. & The exact presence of phrases like "regardless of the way or form" or "in whatever form" leads to \textbf{Entailment} for verbal/oral information, even if "oral" is omitted. \\
\hline
3 & Mandating the return of "all" information with no listed exceptions directly \textbf{contradicts} a claim that a party may retain some. & If the contract requires the return of "all" information and lists no exceptions, the claim that the party may retain some is \textbf{NotMentioned} \\
\end{tabularx}
\end{tcolorbox}

\section{Clustering and Instruction Consolidation}
\label{app:implementation_details}

\subsection{Clustering Implementation details}
\label{app:clustering_details}

We utilized the Gemini Embedding model \cite{lee2025geminiembeddinggeneralizableembeddings} to generate 768-dimensional dense vector representations of all extracted instructions. Clustering was performed using DBSCAN \cite{ester1996density} with a cosine distance metric. We configured the algorithm with a neighborhood threshold of $\epsilon = 0.4$ and a minimum cluster size of $min\_samples=6$. This configuration effectively partitioned the embedding space into 17 distinct semantic clusters for Contract-NLI and 4 broad clusters for StereoSet, while automatically discarding non-generalizable outlier instructions. For the consolidation phase, we employed Gemini 3 Pro using a structured prompt (see Appendix~\ref{app:consolidation_prompt}) to synthesize the raw instructions within each cluster into a unified heuristic.

\subsection{Ablation Study of DBSCAN Clustering Parameters}

\begin{table}[htbp]
\centering
\caption{Sample of Ablation Study of DBSCAN Clustering Parameters for Contract-NLI using Gemini 2 Flash}
\begin{tabular}{cccccc}
\toprule
$\epsilon$ & $min\_samples$ & Total Clusters & Prompt Length (Tokens) & F1 Macro (\%) \\
\midrule
0.2 & 6 & 27 & 6,449 & 79 \\
0.3 & 6 & 22 & 5,580 & 82 \\
\textbf{0.4} & \textbf{6} & \textbf{17} & \textbf{4,630} & \textbf{83} \\
0.5 & 6 & 14 & 4,068 & 80 \\
0.4 & 3 & 24 & 5,914 & 81 \\
0.4 & 9 & 15 & 4,432 & 82 \\
\bottomrule
\end{tabular}
\label{tab:ablation_dbscan}
\end{table}

Table~\ref{tab:ablation_dbscan} shows the chosen configuration ($\epsilon = 0.4$, $min\_samples = 6$)  acting as the optimal balance. It generates 17 distinct clusters with a manageable prompt length of 4,630 tokens, yielding the highest accuracy of 83

\textbf{The Impact of Epsilon ($\epsilon$):}
\begin{itemize}
    \item \textbf{Low Epsilon (0.2):} The algorithm requires vectors to be extremely close to form a cluster. This leads to high fragmentation (27 clusters). The resulting system prompt becomes bloated (4,200 tokens). The model struggles to process these overly specific instructions, causing a drop in F1 score.
    \item \textbf{High Epsilon (0.5):} The algorithm merges distinct reasoning paths together. This collapses the logic into just 14 broad clusters. The prompt is shorter, but the instructions are more generic to handle nuanced edge cases, resulting in the lower F1 score.

\end{itemize}

\textbf{The Impact of Minimum Samples ($min\_samples$):}
\begin{itemize}
    \item \textbf{Low Minimum Samples (3):} Lowering this threshold allows smaller and more niche patterns to form clusters. While it captures more specific rules, it inflates the prompt to 24 clusters and 5,914 tokens. The drop in accuracy suggests the model is starting to overfit to minor variations in the training data.
    \item \textbf{High Minimum Samples (9):}  Raising the threshold forces the algorithm to recognize only highly frequent reasoning patterns. It reduces the cluster count to 15. The prompt loss of some long-tail logic reduces overall performance.

\end{itemize}

\subsection{Instruction Consolidation Prompt}
\label{app:consolidation_prompt}

To synthesize the noisy, redundant micro-instructions within each cluster into a single, high-fidelity heuristic, we employed a Logic consolidation prompt. This prompt forces the model to generalize specific entities while rigorously preserving the causal logic.~\ref{lst:synthesis_prompt}.

\begin{tcolorbox}[
    title=Logic Synthesis Prompt,
    colback=gray!5!white,
    colframe=gray!80!black,
    breakable,
    label={lst:synthesis_prompt}
]
\textbf{System:} You are an expert Logic Synthesizer. \\

\textbf{Task:} You will be given a list of raw "micro-instructions" extracted from different training examples that form a single semantic cluster. Your goal is to:
\begin{enumerate}
    \item \textbf{Identify the Topic:} Provide a short, specific label (2-4 words) that summarizes the legal or logical concept (e.g., "Reverse Engineering", "Audit Rights").
    \item \textbf{Synthesize the Logic:} Consolidate the raw instructions into a single, comprehensive \textbf{Master Instruction} that preserves all critical conditions and keywords.
\end{enumerate}

\textbf{Input:} \\
$[$List of raw instructions$]$ \\

\textbf{Guidelines:}
\begin{enumerate}
    \item \textbf{Generalize:} Remove specific entity names (dates, company names, dollar amounts) unless they are critical thresholds (e.g., "\$100k limit").
    \item \textbf{Preserve Logic:} Ensure the \textit{causal mechanism} (If X then Y) is preserved.
    \item \textbf{Format:} Output a single executable instruction in clear natural language.
\end{enumerate}
\textbf{Output Schema:} \\
\begin{verbatim}
{
  "topic": "Short Topic Name",
  "instruction": "The synthesized master instruction text..."
}
\end{verbatim}
\end{tcolorbox}

\section{Scalability Ablation}

We ran an ablation study on the Contract-NLI dataset by iteratively increasing the training data size to observe its impact on the total number of clusters and the resulting prompt length. With just 1,030 examples, the framework successfully identified the main logical topics, establishing a baseline of 16 clusters. As the dataset size increased to over 7,000 examples, the total number of clusters plateaued at 18, meaning the prompt length remained highly stable. Instead of generating new topics, the additional data refined the existing clusters, which steadily improved the F1 macro score from 0.77 to 0.83 without introducing prompt bloat.

\begin{table}[htbp]
\centering
\caption{Ablation Study of Dataset Size, Prompt Length (Tokens) and F1 score for Contract-NLI using Gemini 2 Flash}
\label{tab:ablation_study}
\begin{tabular}{l c c c c}
\toprule
\textbf{Index} & \textbf{Dataset Size} & \textbf{Total Clusters} & \textbf{Prompt Length (Tokens)} & \textbf{F1 Macro} \\
\midrule
1     & 1,030 & 16 & 4,062 & 0.77 \\
2  & 2,060 & 18 & 4,486 & 0.79 \\
3  & 3,090 & 18 & 4,410 & 0.80 \\
4  & 4,119 & 18 & 4,580 & 0.80 \\
5  & 5,148 & 18 & 4,520 & 0.82 \\
6  & 6,178 & 18 & 4,665 & 0.83 \\
7  & 7,190 & 18 & 4,630 & 0.83\\
\bottomrule
\end{tabular}
\end{table}

\section{Conflict resolution model}

\subsection{Examples of Contract-NLI Conflict resolution instructions}
\label{sec:appendix_conflicts_resolved_examples}

\begin{tcolorbox}[
    title=Hypothesis \#2: Claims the Receiving Party may 'retain' or 'keep' copies (e.g.\, for legal archival purposes).,
    colback=gray!5!white,      
    colframe=gray!80!black,    
    fonttitle=\bfseries,
    breakable                  
]

    \begin{tcblisting}{
        colback=white,           % White background so it stands out from the outer box
        colframe=gray!40!black,  % Slightly lighter border for the inner box
        listing only,
        breakable,
        top=2mm, bottom=2mm, left=2mm, right=2mm % Adds nice inner padding
    }
  **Before Conflict resolution model**
  {
    "condition": "The text mandates return/destruction of 'all' copies and lists NO exceptions",
    "label": "Contradiction"
  },
  {
    "condition": "The text contains an exception clause (e.g., 'subject to,' 'except for,' 'provided that') allowing retention for 'legal,' 'archival,' 'backup,' or 'compliance' purposes",
    "label": "Entailment"
  }
    \end{tcblisting}

    \vspace{0.5em} 

    \begin{tcblisting}{
        colback=white,
        colframe=gray!40!black,
        listing only,
        breakable,
        top=2mm, bottom=2mm, left=2mm, right=2mm
    }
 **After Conflict resolution model**
 {
    "condition": "The text explicitly permits the Receiving Party to 'retain', 'keep', 'store', or 'preserve' copies of Confidential Information (or states return/destruction obligations are 'subject to' such retention) for specific purposes such as 'legal', 'archival', 'backup', 'compliance', 'audit', 'regulatory', 'record-keeping', 'defending claims', or for use by 'legal counsel'.",
    "label": "Entailment"
  },
  {
    "condition": "The text explicitly mandates the 'return', 'destruction', 'deletion', or 'erasure' of 'all' Confidential Information (or 'all copies', 'all records', 'all materials') upon termination or request, AND does not contain exceptions allowing retention for legal/backup/archival purposes (Note: A clause stating items that 'cannot be returned' must be 'destroyed' is considered a mandate for elimination, not retention).",
    "label": "Contradiction"
  },
  {
    "condition": "The text is silent regarding the return or destruction of Confidential Information upon termination/request, or does not explicitly specify whether copies may be retained.",
    "label": "NotMentioned"
  }
\end{tcblisting}
\end{tcolorbox}

\begin{tcolorbox}[
    title=Hypothesis \#6: Claims the party is prohibited from 'reverse engineering\,' 'decompiling\,' or 'disassembling.',
    colback=gray!5!white,      
    colframe=gray!80!black,    
    fonttitle=\bfseries,
    breakable                  
]

    \begin{tcblisting}{
        colback=white,           % White background so it stands out from the outer box
        colframe=gray!40!black,  % Slightly lighter border for the inner box
        listing only,
        breakable,
        top=2mm, bottom=2mm, left=2mm, right=2mm % Adds nice inner padding
    }
  **Before Conflict resolution model**
  {
    "condition": "The text contains the keywords: reverse engineer, decompile, disassemble, reconstruct",
    "label": "Entailment"
  },
  {
    "condition": "These specific keywords are absent (even if the contract prohibits 'unauthorized use' or 'copying')",
    "label": "NotMentioned"
  }
    \end{tcblisting}

    \vspace{0.5em} 

    \begin{tcblisting}{
        colback=white,
        colframe=gray!40!black,
        listing only,
        breakable,
        top=2mm, bottom=2mm, left=2mm, right=2mm
    }
  **After Conflict resolution model**
  {
    "condition": "The text explicitly prohibits the act (keywords: reverse engineer, decompile, disassemble, reconstruct, reverse assemble, reverse compile, discover source code, analyze chemical structure, determine chemical composition, recreate material) without providing an exception that allows the act",
    "label": "Entailment"
  },
  {
    "condition": "The text explicitly permits reverse engineering or related activities under certain conditions or exceptions (e.g., 'except as needed for the project' or 'unless authorized'), which contradicts a blanket 'shall not' hypothesis",
    "label": "Contradiction"
  },
  {
    "condition": "The specific keywords (reverse engineer, decompile, disassemble, reconstruct, analyze chemical structure, recreate) are absent (even if the contract prohibits 'unauthorized use' or 'copying')",
    "label": "NotMentioned"
  }
\end{tcblisting}

\end{tcolorbox}

\section{Model Latency}
\label{sec:model_latency}
Latency of various models monitored over 1000 requests.

\begin{figure}[h!]
    \centering
    \includegraphics[width=0.5\linewidth]{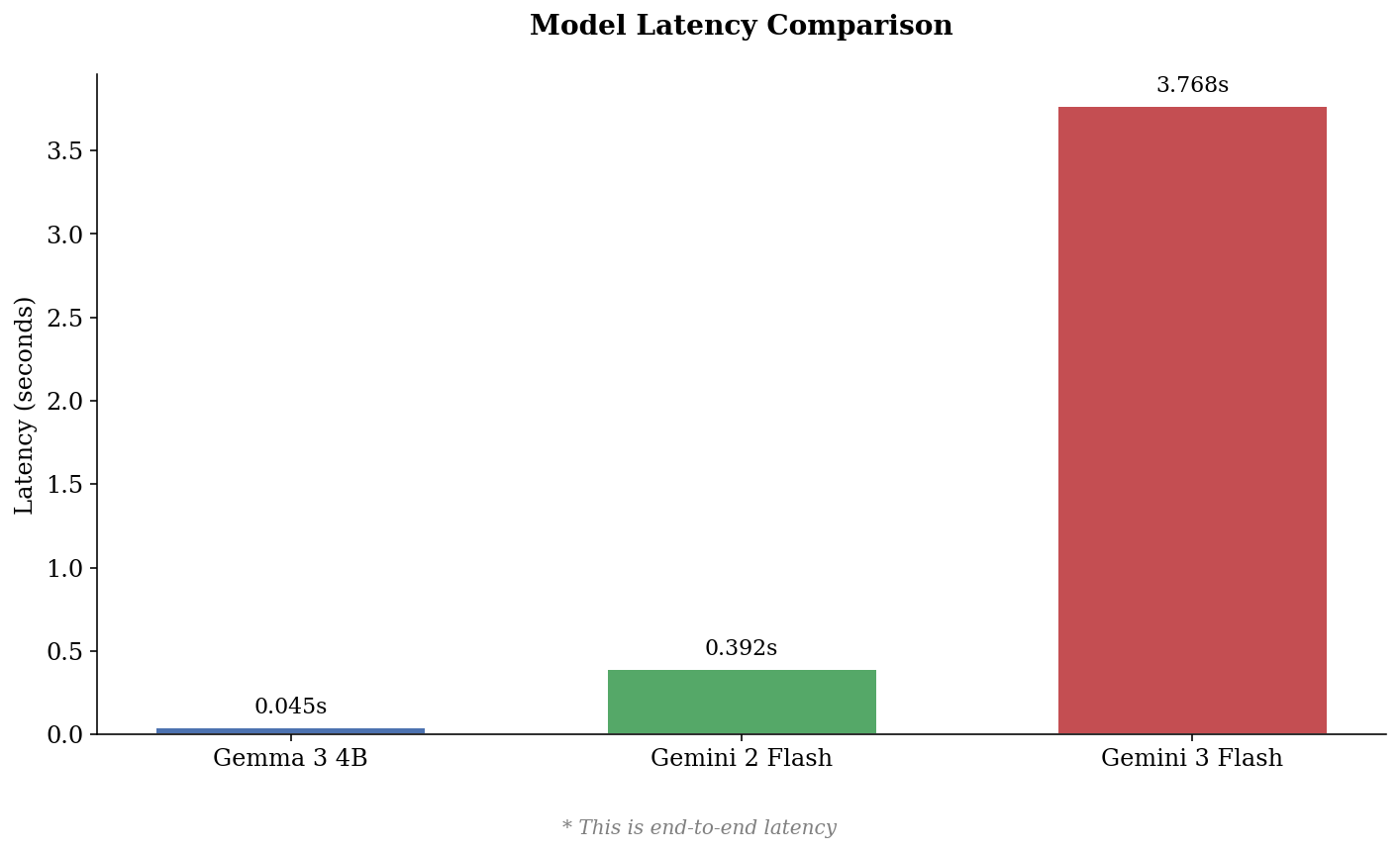}
    \caption{E2E Model Latency between Gemma-3 4B, Gemini 2 Flash, and Gemini 3 Flash}
    \label{fig:placeholder}
\end{figure}

\section{Model Pricing}
\label{sec:model_pricing}
\begin{figure}[h!]
    \centering
    \includegraphics[width=0.5\linewidth]{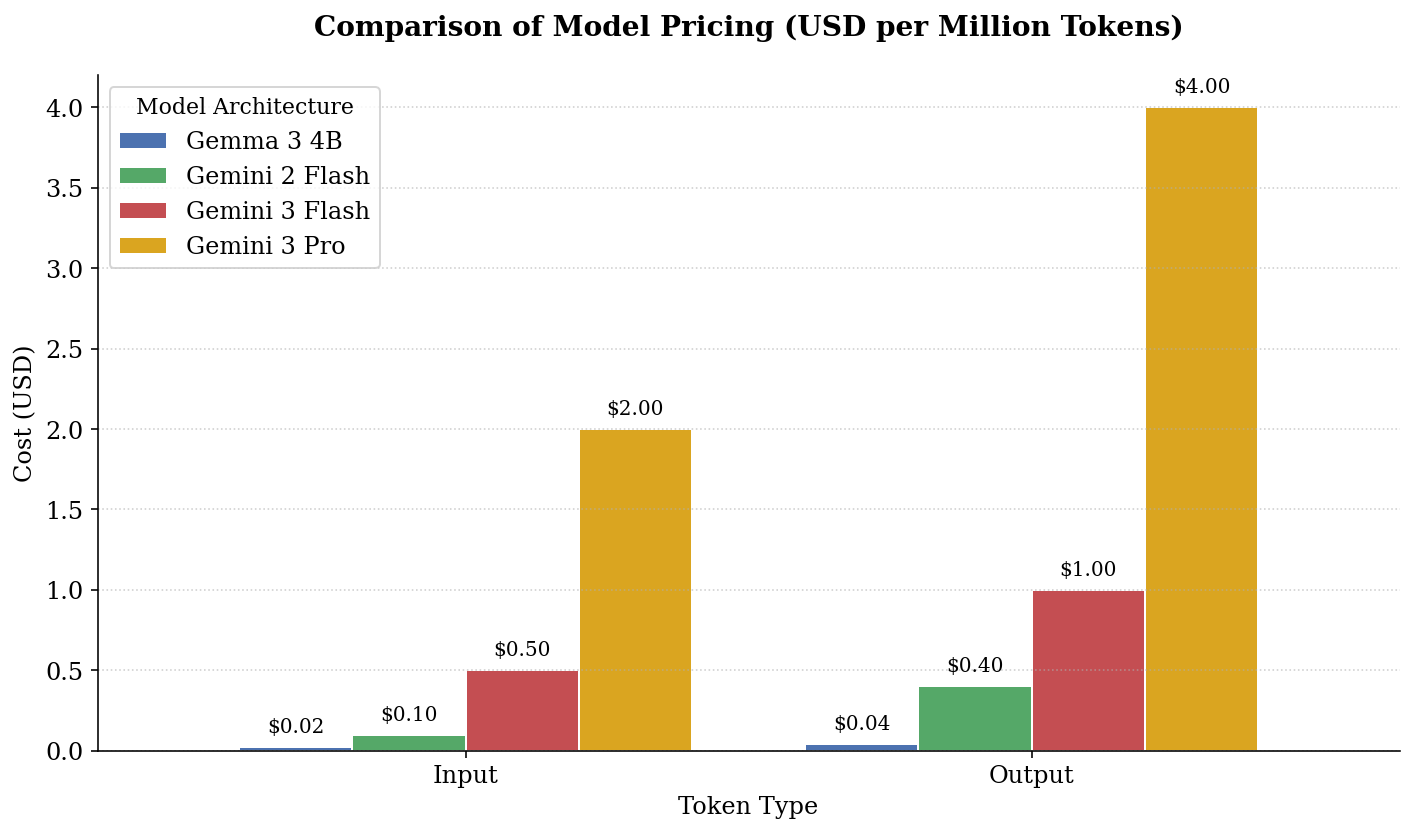}
    \caption{Comparision of Model pricing}
    \label{fig:model_comp}
\end{figure}

\end{document}